\documentclass[xcolor=dvipsnames,journal]{IEEEtran}
\usepackage{graphicx}
\usepackage{cite}
\usepackage{picinpar}
\usepackage{amsmath}
\usepackage{url}
\usepackage{flushend}
\usepackage[utf8]{inputenc}
\usepackage{colortbl}
\usepackage{soul}
\usepackage{multirow}
\usepackage{pifont}
\usepackage{color}
\usepackage{alltt}
\usepackage[hidelinks]{hyperref}
\usepackage{enumerate}
\usepackage[T1]{fontenc}
\usepackage{breakurl}
\usepackage{epstopdf}
\usepackage{pbox}
\usepackage{amsmath,amsfonts}
\usepackage{units}
\usepackage{algorithmic}
\usepackage{array}
\usepackage[caption=false,font=normalsize,labelfont=sf,textfont=sf]{subfig}
\usepackage{textcomp}
\usepackage{cite}
\usepackage[numbers,sort]{natbib}
\usepackage{stfloats}
\usepackage{url,blindtext}
\usepackage{verbatim}
\usepackage{graphicx, epstopdf}
\usepackage{color,colortbl}
\usepackage{lipsum}
\usepackage{capt-of}
\usepackage{mathtools}
\usepackage[numbers,sort]{natbib}
\usepackage{amsmath}
\usepackage{amsfonts}
\usepackage{amssymb}
\usepackage{amsthm}
\usepackage{makecell}
\usepackage{array,booktabs,multirow}
\usepackage{bigdelim}
\usepackage{fixltx2e}
\usepackage{algorithmic}
\usepackage{mdwmath}
\usepackage{mdwtab}
\usepackage{eqparbox}
\usepackage[]{caption}
\usepackage[]{romannum}

\usepackage{booktabs}
\usepackage{fixltx2e}
\usepackage{stfloats}
\usepackage{url}
\usepackage{tikz}

\usepackage{tikz}
\def\checkmark{\tikz\fill[scale=0.4](0,.35) -- (.25,0) -- (1,.7) -- (.25,.15) -- cycle;}

\def\BibTeX{{\rm B\kern-.05em{\sc i\kern-.025em b}\kern-.08em
		T\kern-.1667em\lower.7ex\hbox{E}\kern-.125emX}}

\newcolumntype{x}[1]{!{\centering\arraybackslash\vrule width #1}}

\begin{document}

	\title{Classification of UHF Partial Discharge Signals in Gas-Insulated HVDC Systems Using Neural Networks}
	\title{Generalizable Classification of UHF Partial Discharge Signals in Gas-Insulated HVDC Systems Using Neural Networks }
\author{Steffen~Seitz,~\IEEEmembership{}
	Thomas Götz,~\IEEEmembership{}
	Christopher Lindenberg,~\IEEEmembership{}
	Ronald~Tetzlaff~\IEEEmembership{Senior~Fellow~IEEE}  ~and~ Stephan Schlegel% <-this % stops a space
	\thanks{Steffen Seitz, Christopher Lindenberg and Ronald Tetzlaff are with the Department
		of Electrical and Computer Engineering, Technische Universität Dresden,  01062 Dresden, Germany. Thomas Götz and Stephan Schlegel are with the Department
		Electrical Power Systems and High Voltage Engineering, Technische Universität Dresden,
		E-mail: Steffen.Seitz@tu-dresden.de}}

%	\author{Author~1,~\IEEEmembership{}
	%		Author 2,~\IEEEmembership{}
	%		Author 3~  ~and~ Author 4 \IEEEmembership{Senior~Fellow~IEEE}~  ~and~ Author 5 % <-this % stops a space
	%		\thanks{Author1, Author 3 and Author 4 are with the Department
		%			of XXX in XXX. Author 2 and Author 5 are with the Department of XXX in XXX.
		%			E-mail: Author1@XXX.xx}% <-this % stops a space
	%		\thanks{}% <-this % stops a space
	%		\thanks{Manuscript received XXX, XXX; revised XXX.}}

\maketitle
	\begin{abstract}
		
Undetected partial discharges (PDs) are a safety critical issue in high voltage (HV) gas insulated systems (GIS). While the diagnosis of PDs under AC voltage is well-established, the analysis of PDs under DC voltage remains an active research field. A key focus of these investigations is the classification of different PD sources to enable subsequent sophisticated analysis.

In this paper, we propose and analyze a neural network-based approach for classifying PD signals caused by metallic protrusions and conductive particles on the insulator of HVDC GIS, without relying on  pulse sequence analysis features.  
In contrast to previous approaches, our proposed model can discriminate the studied PD signals obtained at negative and positive potentials, while also generalizing to unseen operating voltage multiples. 
Additionally, we compare the performance of time- and frequency-domain input signals  and explore the impact of different normalization schemes to mitigate the influence of free-space path loss between the sensor and defect location.
\end{abstract}


%\markboth{IEEE TRANSACTIONS ON POWER DELIVERY}%
\markboth{This work has been submitted to the IEEE for possible publication. Copyright may be transferred without notice.}%
{}

\definecolor{limegreen}{rgb}{0.2, 0.8, 0.2}
\definecolor{forestgreen}{rgb}{0.13, 0.55, 0.13}
\definecolor{greenhtml}{rgb}{0.0, 0.5, 0.0}
	\section{Introduction}

	\IEEEPARstart{T}{he}  increasing integration of renewable energy sources into the existing high-voltage grid requires the use of high-voltage direct current (HVDC) systems. This technology is superior to conventional AC technology for transmitting large amounts of power over long distances because of lower losses, and the elimination of reactive power. Apart from high efficiency, the compact installation of high-voltage equipment is also a critical consideration.
	Both requirements, high-efficiency power transmission and space-saving installation, are  met by gas-insulated systems (GIS), which have  been developed for use in transmission systems under AC voltage stress since the  1960s.

\textcolor{black}{A crucial aspect of ensuring fault-free operation in HVDC GIS is the automatic classification of PD-generating defects prior to interpreting the results in terms of fault-free operation \cite{PD_Damage}.}
\begin{figure}[t]
	\centering
	\includegraphics[width=3.5in]{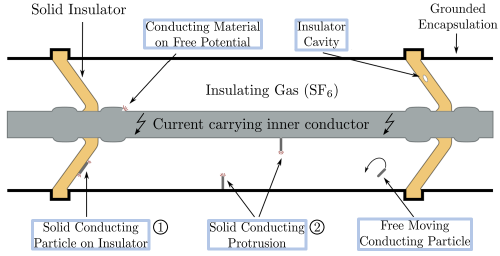}
	\caption{Schematic representation of a HVDC gas-insulated system and related typical partial discharges generating faults. The presented work aims to classificate UHF PD signals caused by particle- \includegraphics[width=0.13in]{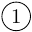} and protrusion-based \includegraphics[width=0.13in]{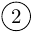} %\circled{2}
		defects. \cite{GoetzDISS}
	}
	\label{fig:GIS}
	
\end{figure} 
\textcolor{black}{  Figure \ref{fig:GIS} provides a visual representation of an HVDC GIS and illustrates common sources of PDs, including solid metallic particles on the insulation, on free potential or freely moving in the insulating gas, and conducting protrusions on the
 encapsulation or conductor \cite{ GoetzDISS}.}
%\textcolor{black}{	
%PRPD features can also be used as input to machine learning methods, enabling PD diagnosis under AC without relying on human expert evaluation \cite{ACPDNeuralNetwork, PRPD_CNN}.}
%\textcolor{black}{
% Recent approaches aim to further improve non-human AC PD classification by  directly utilizing UHF time-domain signals as input to neural networks, including multi-layer perceptrons (MLPs) or convolutional neural networks (CNNs). These studies have shown promising results in terms of improving the accuracy of non-human AC PD classification in GIS \cite{ACPD_CNN, ACCNN}, as well as in related gas insultated lines (GIL) \cite{GIL}.}
\textcolor{black}{ In contrast to conventional AC GIS, the number of these devices in operation under DC stress  is rather low. Thus, the measurement \cite{Cigre}, classification \cite{Pirker}, and physical interpretation \cite{Promotion} of DC PDs continue to be active areas of research. Recently, studies investigating PD development in HVDC GIS indicate that the physical processes responsible for PD formation are the same as compared to those observed under AC voltage stress \cite{GoetsProtrusion}. However, due to the constant electric field, continuous directed movement of charge carriers, and the generation of space and surface charges, the behavior of DC PDs events, such as amplitude and repetition rate, differs significantly from AC PDs \cite{ GoetzDISS}.} 

\textcolor{black}{As a result, the methods and findings related to AC PD classification cannot be directly applied to HVDC GIS.
For example, the  well-established AC GIS PD detection method, which relies on measurements in the ultra-high frequency (UHF) range and  human expert  evaluation of phase-resolved partial discharge (PRPD) plots \cite{UHF, UHF2}, is not suitable for distinguishing DC PD source signals due to the lack of necessary phase information. Consequently, the development and testing of novel DC-specific PD classification methods are essential for ensuring the safety of HVDC GIS.}
\label{relatedwork}
% are considered complex \cite{ Promotion} and
%continues to be actively researched  \cite{Promotion, GoetsProtrusion}. 

\textcolor{black}{The most advanced technique for evaluating and identifying PDs under DC voltage stress is pulse sequence analysis (PSA) followed by the assessment of  patterns by human experts \cite{PirkerPattern}. This method leverages the amplitude and time information of individual PD events in the UHF signal to identify the  underlying defect \cite{NoDI}. However, this approach relies on time-consuming human judgment and it is limited to scenarios with a single PD source. In situations where multiple sources are active, the PD signals overlap, hindering clear identification and necessitating the use of complex techniques for source separation \cite{HochbrueckerFilter, Hochbrueckner}.}
%The work of  \citet{Hochbrueckner} compared different clustering algorithms like histogram thresholding and k-medoids for UHF time-domain DC-PD detection. Despite being able to distinguish two different PD signals, the downside of these approaches is their dependence on prior knowledge about the number of clusters in a signal.

\textcolor{black}{To circumvent the time-intensive process of human evaluation, \citet{Schober} proposed a machine learning-based approach for autonomously classifying DC PD signals in GIS. Their novel approach employed Support Vector Machines and MLP to assess  PSA features derived from UHF measurements. Despite its promising results, this method  inherits the limitations associated with hand-crafted PSA features, such as challenges in classifying signals of multiple active PD sources.} 

\begin{figure}[t]
	\centering
	\includegraphics[width=3.5in]{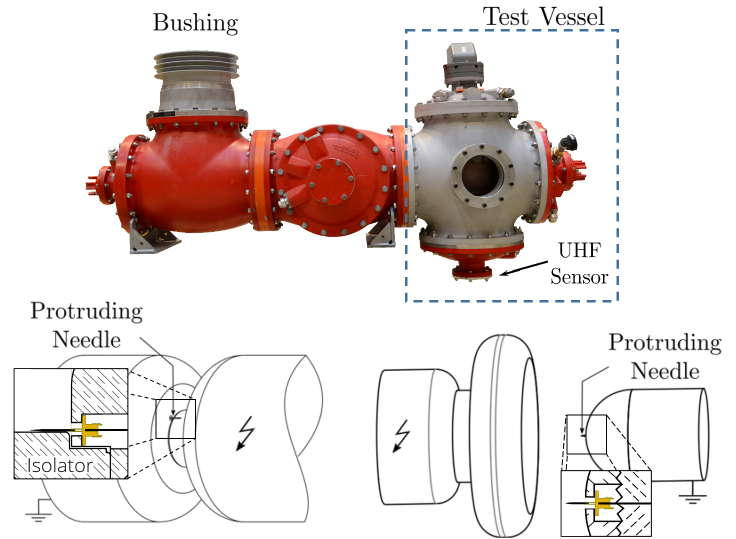}
	\caption{Picture of the experimental setup (top) and three-dimensional schematics of the electrode arrangements used for the emulation of DC-PD at the gas-solid interface (bottom, left) and at a fixed metallic protrusion (bottom, right) \cite{GoetzDISS}.}
	\label{fig:TestSetup}
	
\end{figure} 

\textcolor{black}{Furthermore, evaluation methods based on neural networks have shown promising results in effectively distinguishing multiple active signal sources in similar domains, when directly applied to time-domain measurements \cite{Tenbohlen, Overlap}. Consequently, when these techniques are used for DC GIS PD classification, they could potentially eliminate the need for PSA feature evaluation and source separation techniques.}

\textcolor{black}{Recently, \citet{BeuraCNN} introduced a first neural network-based method for autonomous feature extraction from UHF time-domain measurements. Their study showed results comparable to the pre-existing PSA methods \cite{Schober}, proving the efficiency of this new approach. However, several critical questions that are essential for the practical application of this method remain open.} 	

\textcolor{black}{First, as already illustrated in Figure \ref{fig:GIS}, DC PDs in GIS originate from multiple sources. Among these source types, PDs generated by particles on the insulator surface are particularly significant \cite{NoDI}. However, the existing literature on PD classification using neural networks based on UHF measurements does not provide sufficient evidence regarding the capability to classify this specific fault type.
Moreover, there is a lack of information regarding the classification accuracy of UHF signal-based neural network models specifically for protrusions and insulator particles at both negative and positive polarities of the inception voltage. This information is crucial for a risk assessment of the asset and a successful transfer of the laboratory experiment results to on-site installations \cite{GoetzDISS}.}

\textcolor{black}{In addition, previous studies under AC stress have demonstrated that utilizing fast Fourier transform (FFT) coefficients extracted from UHF signals as training data yields improved classification results in similar AC classification tasks \cite{FFT}. Therefore, it would be advantageous to explore the transferability of these findings to the domain of DC PD classification.}
%Thus, it is important to investigate, if neural network models are able to classify signals from other relevant PD sources in DC-GIS and whether the classification results can be improved by transferring the UHF time-domain signal to the frequency-domain. 

Furthermore, all typical defects of HVDC GIS can experience stress from various multiples of the inception voltage $U_\mathrm{i}$. These individual voltages lead to slightly different discharge patterns \cite{GoetzDISS}. However, the specific  $U_\mathrm{i}$ at the on-site GIS is not known in practice and creating a suitable amount of data for every combination of the defect type and $U_\mathrm{i}$ multiple is impractical for experts. 
Thus, it is essential to investigate the ability of possible DC PD classification models to generalize from laboratory-based signals to measurements based on unseen  $U_\mathrm{i}$ multiples of the GIS.

Another challenge in HVDC GIS arises from the amplitude of UHF PD signals, which is influenced by free-space path loss, particularly when the signal has to cross barrier insulators. Thus, the amplitude is proportional to the distance between the defect location and the sensor. As this distance varies in applications outside laboratory tests, it might be advantageous to generally exclude amplitude-related information during model training by applying different normalization methods. 
\textcolor{black}{In summary, this study contributes to the HVDC GIS PD classification in three significant aspects:}

%\begin{itemize}
%	\item \textcolor{black}{First, we extend the approach of \citet{BeuraCNN} by providing the missing evidence that a neural network-based architecture can effectively classify DC PD time-domain UHF signals originating from  particles on an insulator and  fixed metallic protrusions at both negative and positive polarities of the GIS. Furthermore, we aim to improve the classification performance by incorporating frequency-domain signals and exploring different layer configurations, including the number, ordering, and hyperparameters (stride, kernel size) of our network.}
%	
%	\item \textcolor{black}{ Second, we present the first investigation of an HVDC PD classification model in terms of its ability to classify measurements of a never-trained multiple of the GIS inception voltage $U_\mathrm{i}$. In doing so, we aim to provide insight into the transfer learning ability of the model to generalize from laboratory data to data recorded from on-site installations under unknown DC voltage stress levels.}
%	
%	\item 	\textcolor{black}{Third, we analyze the effect of different normalization methods on the performance of a HVDC PD classification model to mitigate the possible influence of free-space path losses between the sensor- and defect location.}
%\end{itemize}	

\begin{itemize}
	\item \textcolor{black}{First, we extend the approach of \citet{BeuraCNN} by providing the missing evidence that a neural network-based architecture can effectively classify DC PD time-domain UHF signals originating from particles on an insulator and fixed metallic protrusions at both negative and positive DC voltage stress. Furthermore, we aim to improve the classification performance by incorporating frequency-domain signals and exploring different layer configurations, including the number, ordering, and hyperparameters (stride, kernel size) of our network.}
	
	\item \textcolor{black}{Second, we present the first investigation of an HVDC PD classification model in terms of its ability to classify measurements obtained at multiples of the inception voltage $U_\mathrm{i}$ that have never been included in the model training. By doing so, we seek to gain insights into the transfer learning capabilities of the model, allowing for generalization from laboratory data to on-site installations under unknown DC voltage stress levels.}
	
	\item \textcolor{black}{Third, we analyze the impact of different normalization methods on the performance of an HVDC PD classification model to mitigate potential influences of free-space path losses between the sensor and defect locations. This analysis aims to enhance the robustness of the model in real-world scenarios where the distance between the sensor and defect location may vary.}
\end{itemize}

%
% 
%
%
%With a technical point of view the diagnosis system can classify the measured PD signals into different classes and subsequently an identification of the PD sources present within the gas-insulated system using the PSA-method would be possible.

\noindent

\section{Experimental Setup \& Methods}

This study is conducted using UHF DC-PD signals measured under DC voltage stress, as documented in \citet{GoetzDISS} and \citet{GoetsProtrusion}. The measurements were performed in a gas-insulated test setup shown in Figure \ref{fig:TestSetup} (top). Similar to \citet{BeuraCNN}, model electrode arrays are installed inside a test vessel to simulate the typical behaviour of the defects. 
The tests were performed using sulfur-hexafluoride (SF\textsubscript{6}) as the insulation gas at an absolute pressure of 0.5 MPa. 

In the protrusion arrangement (Figure \ref{fig:TestSetup}; bottom, right), the metallic protruding needle has a length of  5\,mm and the needle tip is installed at a distance of $ 55  $ mm from the opposing high-voltage electrode. 
To simulate  PD at the gas-solid interface (Figure \ref{fig:TestSetup}; bottom, left),  a $ 13  $ mm needle  is placed on the surface of an epoxy insulator, typically used in gas-insulated systems. The distance between the needle tip and the opposing high-voltage electrode is 26.7\,mm.
The high DC voltage with positive and negative polarity up to voltages of 250\,kV is applied via the air-SF\textsubscript{6} bushing installed in the centre of the test vessel. PD signals are detected using a standard UHF sensor \cite{Troeger} installed in the test vessel. A 30 dB amplifier is used to increase the signal-to-noise ratio.

The time-dependent PD signals are sampled at $10 $ GS/s using a Teledyne LeCroy WavePro 735 ZiA digital oscilloscope.
Each $i$-th measurement $M_j = \{s_{j,1}, ... , s_{j,u},... ,s_{j,l_s}\}, j\in\{1, \ldots ,N\}$ has a length of $l_\mathrm{s} = 20002$ individual samples $s_{j,u}| s\in \mathbb{R}$, sampled at time step $u$.  The resulting data set contains a total of $N = 33000$ individual PD measurements taken at different multiples of the inception voltage $U_\mathrm{i}$.  
\begin{figure}[t]
	\includegraphics[width=3.5in]{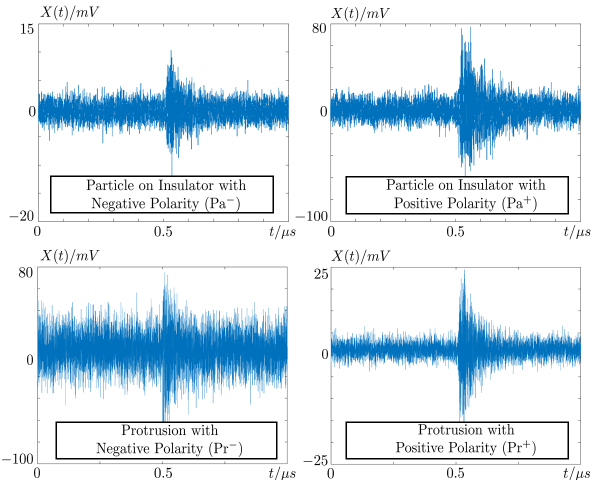}
	%		\caption{UHF time-domain signals obtained from monitoring the DC-PD-generating defects considered in this work. Each measurement is based on either protrusions on the encapsulation (Pr$^-$, Pr$^+$) or particles at the gas-solid interface (Pa$^-$, Pa$^+$) at positive and negative potential of the needle electrode.}
	\caption{Time-domain signal examples of the considered defects. Each PD measurement is based on either particles at the gas-solid interface (Pa$^-$, Pa$^+$) or protrusions (Pr$^-$, Pr$^+$) at negative and positive potential of the needle electrode.}
	\label{fig:Sample}
	
\end{figure}    
A so-called source class label is assigned to each $M_j$ based on the defect type, the polarity of the needle and the $U_\mathrm{i}$ multiple. Note that the $U_\mathrm{i}$-information is only utilized to investigate the generalization capability of the model (Section \ref{General}). In the classification task the model discards this information and learns to predict the so-called output class labels  depicted in Figure \ref{fig:Sample} based on the given source class measurement. Thus, the output class labels are based only on the defect type and needle polarity. The number of  available source class measurements is shown in Table \ref{tab:datasetsize}.
%\textcolor{black}{Note that including data from all common PD-generating defects described in Figure \ref{fig:GIS} in our studies would be beneficial. However, these defects require specific experimental setups, and as a result, measurements for these defects are missing in our dataset \cite{GoetzDISS}}.
%		\begin{table}[h]
	%	\caption{Available measurements in the dataset of \cite{GoetzDISS}. }
	%	\centering
	%	\begin{tabular}{_l^l@{\,}^c@{\,}^l} \toprule
		%	\rowstyle{\head}
		%	Class              & Measurements& &      \\ \midrule
		%	Pa$^-_{1\cdot U_\mathrm{i}}$,          & Linux          & &       \\
		%	Pa$^ +_{1\cdot U_\mathrm{i}}$                 & Linux          & \smash{$\left.\rule{0pt}{4.5ex}\right\rbrace$}& Debian      \\
		%Pr$^-_{2\cdot U_\mathrm{i}}$                & Linux          & &       \\
		%	Datenbanken             & Linux          & \multirow{2}{*}{\smash{$\left.\rule{0pt}{3ex}\right\rbrace$}}& \multirow{2}{*}{Red Hat}     \\
		%	DHCP / DNS              & Linux          & &             \\
		%	LDAP / Domaincontroller & Windows        & & Server 2008 \\ \bottomrule
		%\end{tabular}
		%
		%	\label{tab:datasetsize}
		%\end{table}
				\begin{table}[h]
			\caption{Number of available measurements $M_j$ of each source class in the dataset of \cite{GoetzDISS}. }
			\centering
			\begin{tabular}{c  c|c|c x{2pt}c|c|c}
				\toprule
				&\multicolumn{3}{c}{Negative Polarity}&\multicolumn{3}{c}{Positive Polarity}\\ 
				&\multicolumn{3}{c}{$\overbrace{\hspace{28ex}}$}&\multicolumn{3}{c}{$\overbrace{\hspace{30ex}}$}\\ 
				\parbox[t]{2mm}{\multirow{4}{*}{\rotatebox[origin=c]{90}{Particle}}}\ldelim\{{4}{1mm}&Pa$^-_{1\cdot U_\mathrm{i}}$&Pa$^-_{1.5\cdot U_\mathrm{i}}$&Pa$^-_{3\cdot U_\mathrm{i}}$&Pa$^+_{1\cdot U_\mathrm{i}}$&Pa$^+_{1.25\cdot U_\mathrm{i}}$&Pa$^+_{1.5\cdot U_\mathrm{i}}$ \\
				\cline{2-7}

				&&&&&&\\
				&3500&3500&3500&3500&3500&3500\\ 
				&&&&&&\\
				\bottomrule	
				\toprule	
				
				\parbox[t]{2mm}{\multirow{3}{*}{\rotatebox[origin=c]{90}{Protrusion}}}\ldelim\{{4}{1mm}&Pr$^-_{2\cdot U_\mathrm{i}}$&Pr$^-_{3\cdot U_\mathrm{i}}$&-&Pr$^+_{2\cdot U_\mathrm{i}}$&-&-\\
				\cline{2-7}
				&&&&&&\\
				&4000&4000&-&4000&-&-\\
				&&&&&&\\
				\bottomrule	
			\end{tabular}
			\label{tab:datasetsize}
		\end{table}
	
			In each of our experiments, the data is randomly drawn from these recorded measurements and divided into a training and a test set by allocating 80 \% of the  measurements of each class to the training set $D_\mathrm{train}$ and the remaining 20 \% to the test set $D_\mathrm{test}$ which are normalized using three different methods.
		\begin{figure*}[!t]
			
			\centering{
				\hspace*{0mm}
				\subfloat{\includegraphics[width=6.3in]{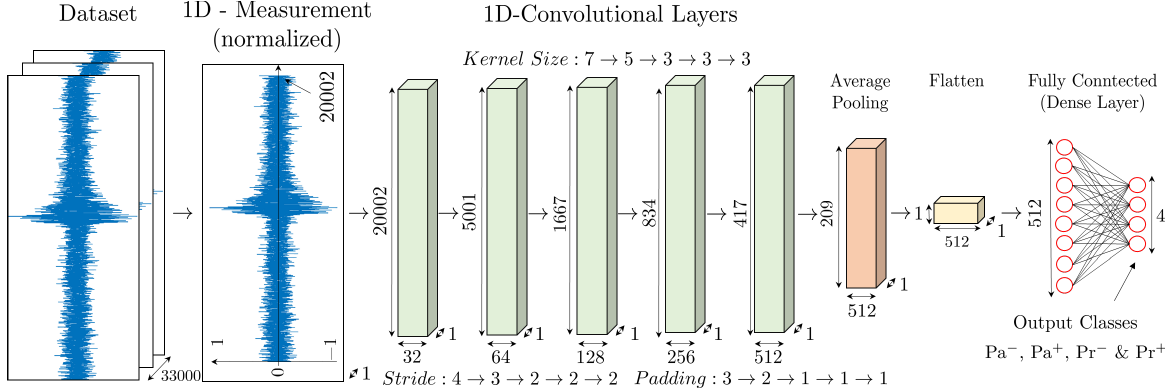}}}
			\hspace*{0mm}
			
			\caption{The proposed model consists of five consecutive 1D-CNN layers with ReLU activation, followed by an average pooling- and a flatten layer. The features are fed into a fully connected multi-layer perceptron. This network has 512 ReLU neurons in the input layer and four output neurons with a softmax activation to assign one of the output classes (Pa$^-$, Pa$^+$, Pr$^-$, Pr$^+$).}
			\label{fig_Network}
		\end{figure*} 
		\subsection{Normalization methods}
		
		The objective of data normalization or min-max scaling is to standardize features to a consistent scale. This typically leads to improved performance and training stability of the model. However, dependent on the choice of the method, its application may affect the information contained in the signal. \textcolor{black}{ Normalization can be applied in three different ways:}
		
		\subsubsection{Trainset normalization (Tr) }

		In trainset normalization,  every sample in each  measurement $M_j$ within $D_\mathrm{train}$ and $D_\mathrm{test}$ is normalized between $-1$ and $+1$ according to:
		
		\begin{equation}
			\bar{s}_{j,u} = \frac{s_{j,u} - \text{Min}(D_\mathrm{train})}{\text{Max}(D_\mathrm{train}) - \text{Min}(D_\mathrm{train})}	.
			%y_k = \text{ReLu}\Big(\sum_{i=0}^{k_s}   \boldsymbol{\theta_{k}}\cdot\mathbf{W_k} + b  \Big)
			\label{train}
		\end{equation}
		
		Min/Max$(\cdot)$ are operations that return the min-/maximum value of any  individual measurement within the dataset. As shown in equation \eqref{train}, trainset normalization scales each sample in each measurement with respect to the maximum and minimum sample value of all individual measurements across all classes in the dataset. Thus, this method preserves the amplitude information for each measurement in the dataset. However, due to the previously stated free-space path loss problem, it might be advantageous to generally exclude amplitude-related information during model training. 
		\subsubsection{Class normalization (Cl)}
		Class normalization scales each sample within a measurement relative to the maximum and minimum sample values of all measurements belonging to the corresponding source class from Table \ref{tab:datasetsize}. In this method, the amplitude-related information is preserved within samples of one class, while it is hidden between measurements of different classes.

		\subsubsection{Measurement normalization (Me)}
		In measurement normalization, each sample in a measurement $M_j$   is scaled based on the maximum and minimum sample values within that specific measurement period in the dataset. In this case, the model can no longer rely on amplitude-related features to classify different samples of class in the dataset. 
		
		\subsection{Classification model}
		\label{cm}
		
		In our experiments, the normalized UHF amplitude signal measurements and their respective FFT coefficients are separately used to train the neural networks-based model as depicted in Figure \ref{fig_Network}. These networks use sample measurements and their assigned output class labels in $D_\mathrm{train}$ to learn their weighting and bias parameters and ultimately predict the correct output class label.
		
		%	As described before, the output class label does not contain any information about the $U_i$. The network therefore learns to assign measurements from several Ui-multipls to a single output class.
		
		 %The network is trained with 15 different weight initializations and the received classification accuracy for each class is averaged.
		The architecture in our work  is based on 1D-convolutional \cite{ Seitz2} and perceptron layers \cite{ Seitz}. We tested different numbers and arrangements of these layer types in conjunction with different activation functions such as ReLU, Sigmoid, and Tanh to achieve an improved classification result. For brevity, only the final model structure is given. 
		This structure and the corresponding activation functions were empirically determined with respect to maximizing classification performance. In the final architecture, the 1D-convolutional layers perform sequential, discrete convolutions between a filter kernel and the data. The extracted features are processed by a ReLU activation function after each layer followed by an average pooling layer and a flattening layer.
		The data is further processed by the perceptron layer with 512 neurons and a ReLU activation function, which maps the extracted information to the output classes depicted in Figure \ref{fig:Sample}. 
				 %\cite{Gradient}.
			 
		 The model parameter adjustment is performed using backpropagation and the optimal parameter set is determined by optimizing a cross-entropy loss using ADAM optimization \cite{kingma2017adam}. The initial learning rate of 0.0001 and a batch size of 64 were determined through a random search.
		In each of our experiments, the network is initialized with 15 different weight initialization seeds to account for the effects of a non-optimal start to the optimization process. The number of individual seeds was chosen to represent a reasonable trade-off between the statistical robustness of the result and the computational time of the model. 
		%\textcolor{black}{Each run on an NVIDIA A100 GPU took approximately 24 minutes of training time.}
		\textcolor{black}{	The performance of each trained model is evaluated using unseen data from $D_\mathrm{test}$. In addition, similar to k-fold cross-validation, a different set of measurements from the limited experimental data are randomly assigned to $D_\mathrm{train}$ and $D_\mathrm{test}$ in each of the 15 training and testing procedures. The obtained classification results are  averaged to determine a more reliable  true positive rate $A$ of the proposed architecture for each defect class.} Similarly, the true negative, false negative and false positive rates of the model are determined for all other output classes. %The results are presented in the form of a confusion matrix.% \cite{ConfusionMatrix}.
		
		\section{Experiments \& Results}
		\subsection{Time- and frequency-domain classification results}
		\label{classification results}
		
		In this section, we analyze the capability of the proposed architecture to classify unseen measurements of PD generated by a fixed protrusion or a particle adhering to the gas-solid interface.
		To accomplish this, we utilize UHF signals measured in either the time-domain or frequency-domain at a specific $U_\mathrm{i}$ multiple for each output class.
		In this initial experiment, the baseline dataset D comprises measurements of Pa$^-_{1\cdot U_\mathrm{i}}$, Pa$^+_{1\cdot U_\mathrm{i}}$, Pr$^-_{2\cdot U_\mathrm{i}}$, and  Pr$^+_{2\cdot U_\mathrm{i}}$, which are then divided into $D_\mathrm{train}$ and $D_\mathrm{test}$.
	    However, while it would be beneficial to classify the output class based on data measured at the same constant multiple of $U_\mathrm{i}$, for technical reasons the available dataset only contained an unbalanced measurement distribution \cite{GoetzDISS}. As a result, there is an unequal distribution of measurements recorded at the same $U_\mathrm{i}$ multiples for each output class, as shown in Table \ref{tab:datasetsize}.

%		\begin{figure}[b]
%			\centering
%			\includegraphics[width=3.5in]{img/ACCVierklassen.pdf}
%			\caption{Model confusion matrix ($\bar{A}_{\text{trainset}} =0.9653$)  obtained by utilizing  Pa$^-_{1\cdot U_\mathrm{i}}$, Pa$^+_{1\cdot U_\mathrm{i}}$, Pr$^-_{2\cdot U_\mathrm{i}}$ and Pr$^+_{2\cdot U_\mathrm{i}}$ time-domain measurements,  preprocessed by Trainset Normalization.}
%			\label{fig:ACCvierklassen}
%		\end{figure}    
		 
		 \begin{figure}[h]
		 	\centering
		 	\includegraphics[width=3.5in]{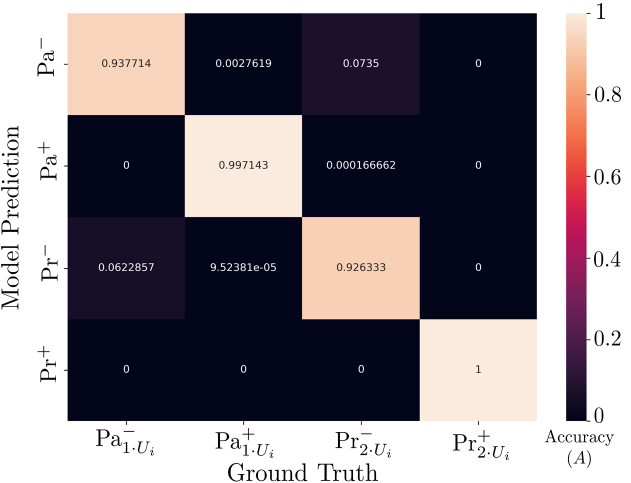}
		 	\caption{Model confusion matrix ($\bar{A}_{\text{trainset}} =0.9653$)  obtained by utilizing  Pa$^-_{1\cdot U_\mathrm{i}}$, Pa$^+_{1\cdot U_\mathrm{i}}$, Pr$^-_{2\cdot U_\mathrm{i}}$ and Pr$^+_{2\cdot U_\mathrm{i}}$ time-domain measurements,  preprocessed by trainset normalization.}
		 	\label{fig:ACCvierklassen}
		 \end{figure}    
		 
		Figure \ref{fig:ACCvierklassen} illustrates the confusion matrix of the experiment involving the proposed output classes
		Pa$^-_{1\cdot U_\mathrm{i}}$, Pa$^+_{1\cdot U_\mathrm{i}}$, Pr$^-_{2\cdot U_\mathrm{i}}$ and Pr$^+_{2\cdot U_\mathrm{i}}$.
		The model achieves a near-perfect true positive rate $A$ for UHF signals resulting from positive particle  (Pa$^+$) or positive protrusion  (Pr$^+$) defects. 
		For the remaining two output classes, the model obtains only a slightly worse result. In particular, it exhibits a nearly symmetrical confusion between measurements associated with negative particles (Pa$^-$) and negative protrusions (Pr$^-$). As mentioned in Section \ref{cm}, the model in this experiment was trained separately with 15 different weight initialization seeds. Thus, the presented classification rates of each class in the confusion matrix are the average across all individual runs. 
		The overall performance of the architecture is represented by the average true positive rate $\bar{A}$. It is determined by averaging the individual true positives of all output classes in the confusion matrix. In Figure \ref{fig:ACCvierklassen},  these values are indicated by the diagonal elements of the matrix, representing the true positive rate of correctly classified PD samples for each class, while the non-diagonal elements represent the false positive and true negative rates when comparing the predicted class against the ground truth.  
		\begin{table}[h]
			\caption{Average true positive classification rates $\bar{A}$ for different normalization strategies and input datatype.  }
			\centering
			\begin{tabular}{c|ccc}
				\toprule
				Input&\multicolumn{3}{c}{Normalization Method}\\
				Datatype & Measurement (Me) & Class (Cl)  & Trainset (Tr)  \\
				\hline
				\\
				
				Time-Domain (TD)  &  \textbf{0.9977} & 0.9932 &  0.9653 \\\\
				Freq.-Domain (FFT)  &  \textbf{0.9983}& 0.9975& 0.8048 \\
				\bottomrule
			\end{tabular}
			\label{fig:AccNormalization}
		\end{table}
	
			As summarized in Table \ref{fig:AccNormalization}, the trained model achieves an $\bar{A}_{\text{Tr}} =0.9653$ on the trainset normalized measurements in $D_\mathrm{test}$. 
		The performance on time-domain test data is generally better if the amplitude information is at least partially- ($\bar{A}_{\text{Cl, TD}} =0.9932$), or completely- $(\bar{A}_{\text{Me, TD}} =0.9977$) hidden after normalization. If the time-domain input data is converted to the frequency-domain instead, the classification rate based on the trainset normalized FFT coefficients only achieves an average of $\bar{A}_{\text{Tr, FFT}} = 0.8048$. Thus, if trainset normalization is selected and the model generalization to unseen $U_\mathrm{i}$ multiples is not required, time-domain data should be used to train the classifier. 
		Analog to the time-domain signal, the FFT-based true positive rate is higher for class- and measurement normalization. Classifying PDs from the FFT coefficiants of the UHF signal resulted in the highest classification performance in our experiments, with $\bar{A}_{\text{Me, FFT}}=0.9983 $. Therefore, the minor difference in performance compared to using the time-domain data hardly justifies the computational cost of converting every measurement to the frequency-domain. Thus, the following experiment relies solely on the time-domain data.

		\subsection{Transfer learning experiment results (generalization to an unseen multiple of $U_\mathrm{i}$)}
		\label{General}
		 
		As previously mentioned in Section \ref{relatedwork}, it is important to examine PD classification models in terms of  their generalization to measurements of defects recorded at unknown multiples of the GIS inception voltage ($U_\mathrm{i}$).  
		Therefore, each PD measurement is assigned a source class based on the defect type and the inception voltage determined at the time of recording. Note that the exact value of $U_\mathrm{i}$ does not need to be further specified, since the approach of this work is centered around the idea of using laboratory data at specific $U_\mathrm{i}$ multiples to train a model that generalizes to data from unknown $U_\mathrm{i}$ multiples at the on-site GIS.
		
		In the first part of our experiment, we evaluate the generalization performance based on a transfer learning framework, where we pre-train our model based on the data of the same source classes used in Section \ref{classification results} (Pa$^-_{1\cdot U_\mathrm{i}}$, Pa$^+_{1\cdot U_\mathrm{i}}$, Pr$^-_{2\cdot U_\mathrm{i}}$ and Pr$^+_{2\cdot U_\mathrm{i}}$), to create a baseline for our further studies. Note, that analog to the previous experiment, any measurement is normalized by either trainset-, class- or measurement normalization.
		After the training process, the generalization performance of the baseline model is assessed by monitoring the model generalization rate $G$, which is equal to the true positive rate of the model on the untrained Pa$^+_{1.5\cdot U_\mathrm{i}}$ measurements in the test set, i.e. $G:= A_{}(\text{Pa}^+_{1.5\cdot U_\mathrm{i}}$). Note, that the data of Pa$^+_{1.5\cdot U_\mathrm{i}}$ is explicitly excluded from the training set in any of the further experiments. herefore, the described measurements are used only at inference time to test the model's generalization to an unseen source class. In this setting, the model can only rely on the knowledge learned from the source class data available in $D_\mathrm{train}$ to classify any of the Pa$^+_{1.5\cdot U_\mathrm{i}}$ measurements in $D_\mathrm{test}$. 
		 
		When trainset normalization (Tr) is used to normalize the baseline dataset (Base), the model achieved a generalization rate of $G_\mathrm{Tr,Base}=0.6801$. With class normalization (Cl), the model achieved a significantly lower result of $G_\mathrm{Cl,Base}=0.4010$, while with measurement normalization, the generalization rate of $G_\mathrm{Me,Base}=0.2085$ did not outperform a random classifier in classifying our four output classes   ($G_\mathrm{r}=0.25$). A possible explanation could be the lack of amplitude-related information after data preprocessing with these normalization methods.
		To investigate this issue further, we refine our basic model by gradually adding measurement data from a previously withheld source class to the trainset $D_\mathrm{train}$. The extended training set is used to train the model from scratch. In parallel, the generalization rate of the model is determined after each addition, analogous to the baseline experiment.
        The sequential training was performed for two different consecutive sequence orders ("Order 1" and "Order 2") to rule out any effects of adding a single record to $D_\mathrm{train}$.

		\begin{figure}[t]
	\centering
	\includegraphics[width=3.5in]{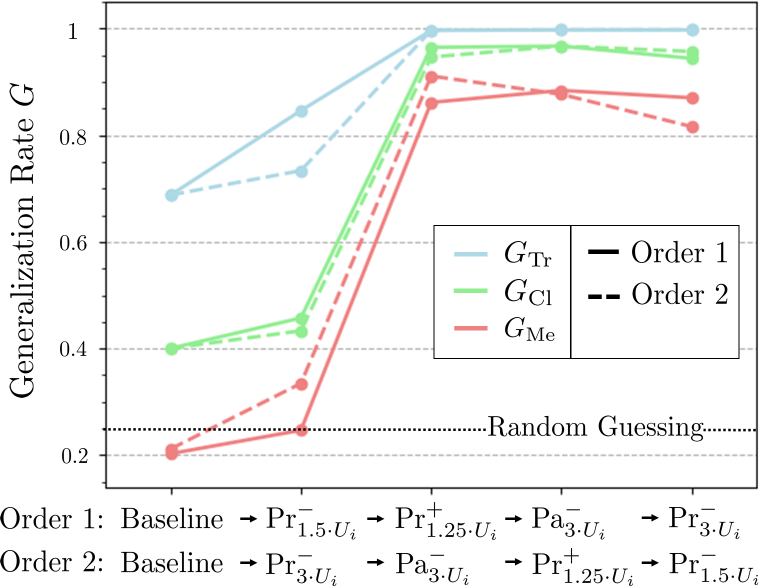}
	\caption{Model generalization rate $G$ on the untrained $U_\mathrm{i}$ multiplier Pa$^+_{1.5\cdot U_\mathrm{i}}$ when additional data from  a specific previously withheld source classes is included in the $D_\mathrm{train}$. The issue is investigated for trainset (Tr), class (Cl), and measurement (Me) normalization and two different sequential training orders.}
	\label{fig:generalisierung}
\end{figure}
The results of this experiment shown in Figure \ref{fig:generalisierung} indicate, that the model achieves a generalization rate that is reliably above the baseline results, regardless of the normalization type and sequential ordering, after data from the first additional $U_\mathrm{i}$ multiple is added to the baseline dataset.
After training the model with data from all available source classes in Table \ref{tab:datasetsize} (except for the $\text{Pa}^+_{1.5\cdot U_\mathrm{i}}$ data), a generalization performance is achieved that far exceeds random guessing. Analog to training with a single additional source class, this result is independent of the type of normalization and the sequential order.
		In contrast to the previous experiment from Table \ref{fig:AccNormalization}, trainset normalization performs best in this transfer learning setting. The highest generalization rate with trainset normalization was achieved after training with order 1 (O1), resulting in $G_{\text{Tr, O1}}=0.9982$. 
		In comparison, the highest generalization rate of class normalization at the end of order 2 (O2) achieved $G_{\text{Cl, O2}}=0.9578$. The highest measurement normalization based  approach only achieves  $G_{\text{Me, O1}}=0.8672$ at the end of order 1. As for the baseline dataset, a possible explanation could be the lack of amplitude-related information after data preprocessing with these normalization methods. 
		In practice, the model is expected to generalize from laboratory measurements to data of a monitored GIS with an unknown $U_\mathrm{i}$ multiple. Therefore, assuming negligible effects of free-space path loss, it is recommended to select normalization methods that preserve the amplitude information in the data.
				\begin{figure}[t]
			\centering
			\includegraphics[width=3.5in]{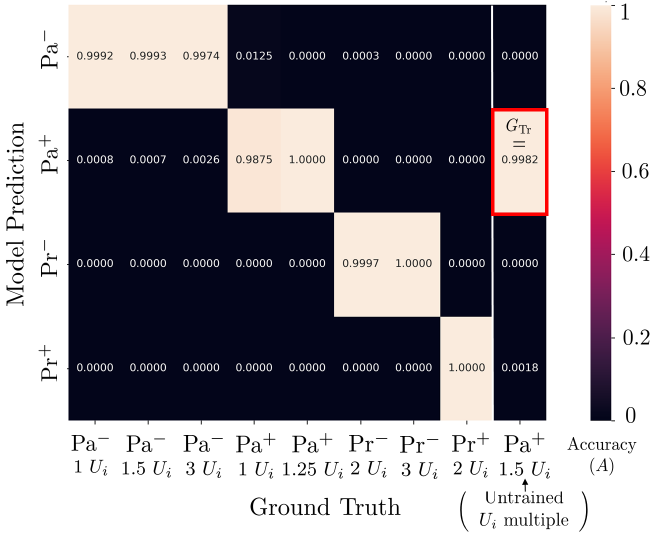}
			\caption{Confusion matrix and generalization rate $G_\mathrm{Full}$ (red) of the proposed model. The network was trained using the full trainset-normalized measurements of all available classes from Table \ref*{tab:datasetsize} (except $\text{Pa}^+_{1.5\cdot U_\mathrm{i}}$), using the train sequence of order 1.    }
			\label{fig:generalisierungResult}
			
		\end{figure} 
	
	Similar to Section \ref{classification results}, it is crucial to evaluate model performance concerning the remaining source class measurements from  Table \ref{tab:datasetsize} while assessing generalization on the withheld $\text{Pa}^+_{1.5\cdot U_\mathrm{i}}$ data. 
	When measurements of all source classes from Table \ref{tab:datasetsize} (exept $\text{Pa}^+_{1.5\cdot U_\mathrm{i}}$) are included in $D_\mathrm{train}$, the model  classifies almost all test set measurements correctly. This is illustrated by the confusion matrix in Figure \ref{fig:generalisierungResult}. For the sake of brevity, all combinations of normalization methods and training routines are not shown, as they achieved comparable results. In the depicted configuration, the average true positive rate over the available classes in the training set is $\bar{A}_{\text{Tr, O1}}=0.9979 $. 
	When this score is averaged with the associated generalization rate  $(G_{\text{Tr, O1}}=0.9982) $, the model attains an average final true positive rate of $\tilde{A}_{\text{Tr, O1}}=0.9977 $ for all measurements across all classes in our experiment.
	
%	 In this context the Pa$^+_{1.5\cdot U_\mathrm{i}}$ measurements excluded from the training process $(G_{\text{Full}}=0.9987) $. If both of these results are averaged the model achieves an average final true positive rate of $\tilde{A}_{\text{Full}}=0.9977 $. 
		\subsection{Comparison to other methods}
\textcolor{black}{ Compared to other methods, our model classifies HVDC PDs originating from metallic particles on the gas-solid interface (insulator) ($\text{Pa}$) of a GIS with an average accuracy of 99.67\%, which has not been reported in any other non-PSA-based DC PD classification study. In addition, our refined CNN/MLP-based architecture achieves an 1.29\% higher average performance than the previous best model in the protrusion ($\text{Pr}$) detection task \cite{BeuraCNN}. The study of \citet{Schober} previously studied Pr and Pa classification, however their PSA-based method did not report individual classification rates for these PD types. In addition, the number of measurements available for each class is very limited, making any comparative analysis on their work even more difficult.}
\textcolor{black}{Furthermore, our model is the first to differentiate between negative and positive polarities of $\text{Pr}^\pm$ and $\text{Pa}^\pm$ measurements, while achieving previously unreported generalization capabilities to untrained $U_\mathrm{i}$ multiples.}
\textcolor{black}{Moreover, we are the first to investigate the effect of normalization on the classification performance.}
\textcolor{black}{The main contributions of our work  to the field are summarized in Table \ref{fig:RelatedWork}.}
		\begin{table}[h]
			\caption{\textcolor{black}{Comparison of the presented work to related non-expert-based  PD classification approaches in HVDC GIS.}  }
			\centering
			\begin{tabular}{c|cccc}
				\toprule
				 
				Category & \textbf{Our work} & \cite{BeuraCNN}  &  \cite{Schober}&  \\
				\hline
				\\
				
				\makecell{ $\text{Pr}$ classification \\ accuracy:   }    &  \textbf{99.99\%} & 98.23\% & \makecell{not\\ reported} &  \\\\
				\makecell{ $\text{Pa}$ classification \\ accuracy:   }    &  \textbf{99.67\%} & \makecell{ not inves- \\ tigated (n.i.)    } & \makecell{ not \\ reported   }     &  \\\\
				\makecell{ Average per class  \\ test set size:   }   &  733 & 461 & 83 &\\\\
				\makecell{No PSA\\ features required:   }    &  \checkmark & \checkmark& \ding{55} &    \\\\
				\makecell{Defect polarity\\  studied  on: }    &   \makecell{   protrusion  \& solid\\  particle on insulator } & \makecell{   particle on\\ free potential } & n.i.  &\\\\
				\makecell{ Normalization  \\ influence   }   &  \checkmark &  n.i. &  n.i. &\\\\
				\makecell{ Generalization \\ to  $U_\mathrm{i}$ multiples:  }   &  \checkmark & \makecell{  n.i.   }  & \makecell{  n.i.  }  &\\\\

				\bottomrule
			\end{tabular}
			\label{fig:RelatedWork}
		\end{table}
\textcolor{black}{ Note that we excluded approaches based on conventional PSA  \cite{PirkerPattern, NoDI} and PDs under AC \cite{ACPD_CNN, ACCNN, ACPDNeuralNetwork,Seitz2} from this comparison, as they either depend on time consuming human expert evaluation or rely on measurements of related but different physical phenomena.}

%  the proposed model is able to generalize from $U_\mathrm{i}$ data of trained source classes to measurements of untrained $U_\mathrm{i}$-multiples, given sufficient training data. Moreover, the model additionally achieves a good classification result on unseen measurements of the trained source class data in our laboratory setting. Additionaly, it seems to be a promising approach for generalizing from laboratory data to on-site measurements.
		
\section{Conclusion}

\textcolor{black}{
	In this paper, we evaluate the performance and generalizability of a neural network-based method for the classification of DC partial discharges (PDs) at both negative and positive polarity. The ultra-high frequency (UHF) time- and frequency-domain PD signals are generated by conductive protrusions and particles on an insulator.}
	Our study extends the literature on HVDC GIS PD classification in multiple aspects:
	\begin{itemize}
		\item \textcolor{black}{	First, we report the first classification of UHF signals originating from particles adhering to the gas-solid interface (insulators) with an accuracy of $99.67$\%, without the need for pulse sequence analysis (PSA). Additionally, our model outperforms all previously reported methods by $1.29$\% on the UHF protrusion signal classification task, achieving an accuracy of $99.99$\%. The use of frequency-domain signals achieved only a negligible average performance advantage of $0.04$\% compared to the UHF time-domain signal classification model result which hardly justifies the computational cost of converting each measurement to the frequency-domain.  }
		
		\item \textcolor{black}{ Second,  we are the first to investigate the transfer learning ability of a UHF signal-based neural network PD classification model to generalize to measurements recorded under unknown DC voltage stress levels. Regardless of the training order, our model correctly classifies $99.87$\% of the $U_\mathrm{i}$ multiple signals not included in the train set as well as $99.77$\% of all available PD  test set measurements.}
		
		\item 	\textcolor{black}{Third,  we analyze the effect of preprocessing, including  sample-, class-, and training set normalization on the performance of a PD classification model in HVDC GIS to mitigate the possible influence of free-space path losses between the sensor- and defect location. Our results show that methods partially eliminating amplitude information perform better in PD classification experiments, while amplitude-preserving methods slightly outperform in the $U_\mathrm{i}$ multiple transfer learning task.}
	\end{itemize}
	
In future work, the proposed approach should be applied to measurements of the remaining PD types in HVDC GIS such as insulator cavitys, free-moving particles, and material on free potential. We also suggest considering different sensor positions, electrode geometries, and spacing to enhance the methodology. Furthermore, it would be important to study the transfer learning capability on data from  additional operating voltages to facilitate the practical application of the method.

% the proposed model is able to generalize from $U_\mathrm{i}$ data of trained source classes to measurements of untrained $U_\mathrm{i}$-multiples, given sufficient training data. Moreover, the model additionally achieves a good classification result on unseen measurements of the trained source class data in our laboratory setting. The presented  neural network-based concept, therefore, seems to be a suitable method to classify UHF signal of PD generated by fixed protrusions and metallic particles adhering to the gas-solid interface in DC-GIS. Additionaly, it seems to be a promising approach for generalizing from laboratory data to on-site measurements.

\section*{Acknowledgments}

This research was partly funded by the German Research Foundation (DFG, project number 379542208) and the German Federal Ministry of Economic Affairs and Climate Action (BMWK, reference: KK5056101KA0).
The authors  acknowledge the computing time through the ZIH at TU Dresden and the  feedback from  Thomas Linde and Carsten Knoll.

\bibliographystyle{Bibliography/IEEEtranTIE}
\bibliography{bibo}{}

% References

\begin{IEEEbiography}[{\includegraphics[width=1in,clip,keepaspectratio]{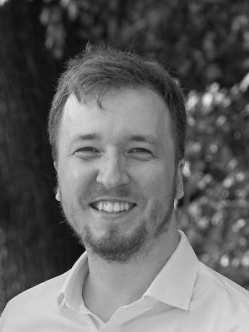}}]{Steffen Seitz}
	received his Diploma (Dipl.-Ing.)
	degree in electrical engineering from
	the Technische Universität Dresden in Germany in 2016. He is currently advancing his academic career as a Ph.D. candidate and research associate at the Chair of Fundamentals of Electrical Engineering, under the guidance of Ronald Tetzlaff. His interests include neural network-based condition monitoring and explainable artificial intelligence and explainable artificial intelligence, with a specific focus on enhancing the understandability of model reasoning in predictive maintenance.\\
\end{IEEEbiography}

\begin{IEEEbiography}[{\includegraphics[width=1in,clip,keepaspectratio]{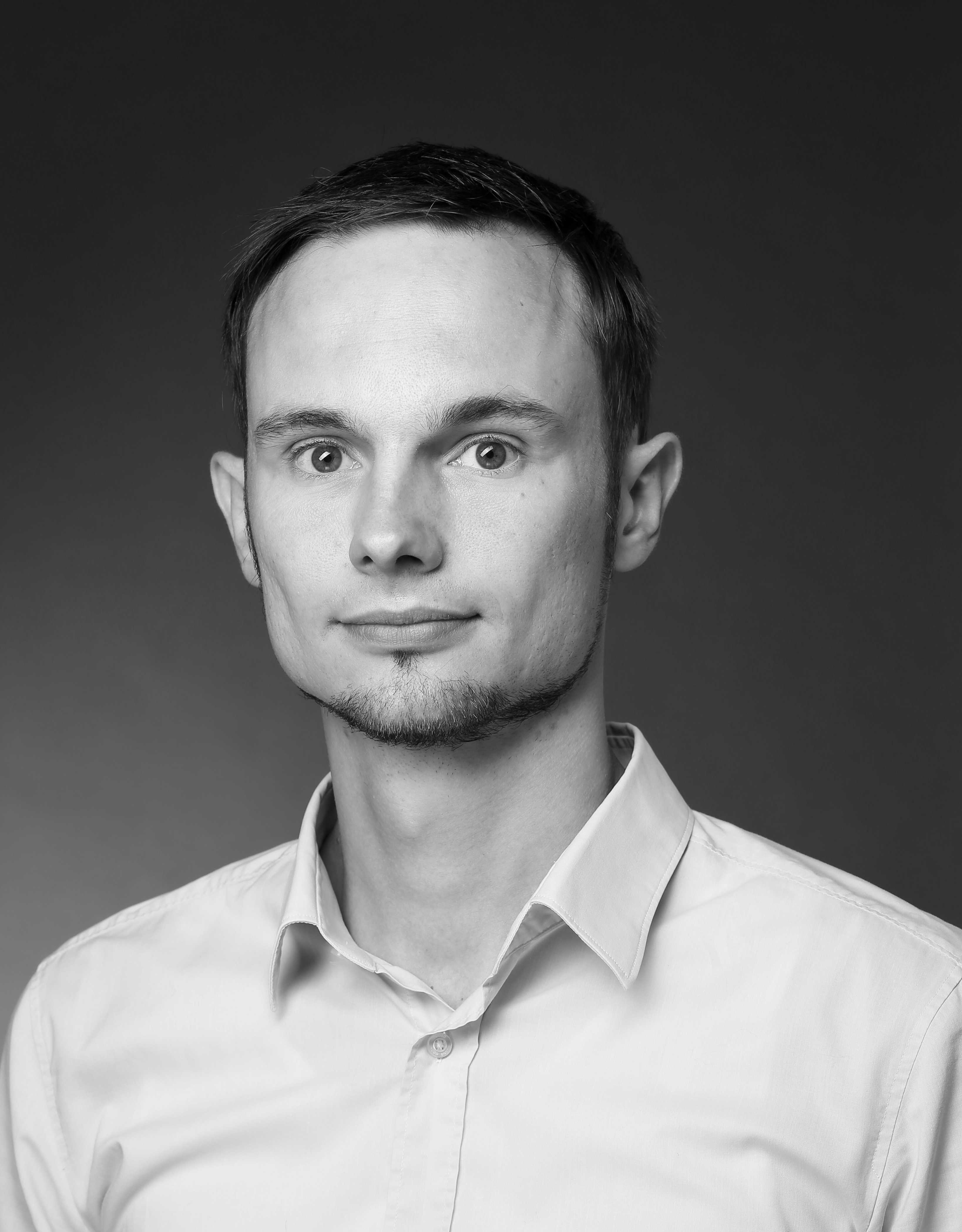}}]{Thomas Götz}
 received the Diploma degree in electrical engineering in 2015 and the Ph.D. degree in high voltage engineering in 2022 from the Technische Universität Dresden.  From 2015 to 2022 he was a research associate at the Chair of High Voltage and High Current Engineering, where he headed the research group High Voltage Engineering in 2022 and 2023. He is a member of the CIGRE WG D1.63 “Partial discharge detection under DC voltage stress”. One of his journal publications was awarded a national literature award.
\end{IEEEbiography}
\begin{IEEEbiography}[{\includegraphics[width=1in,clip,keepaspectratio]{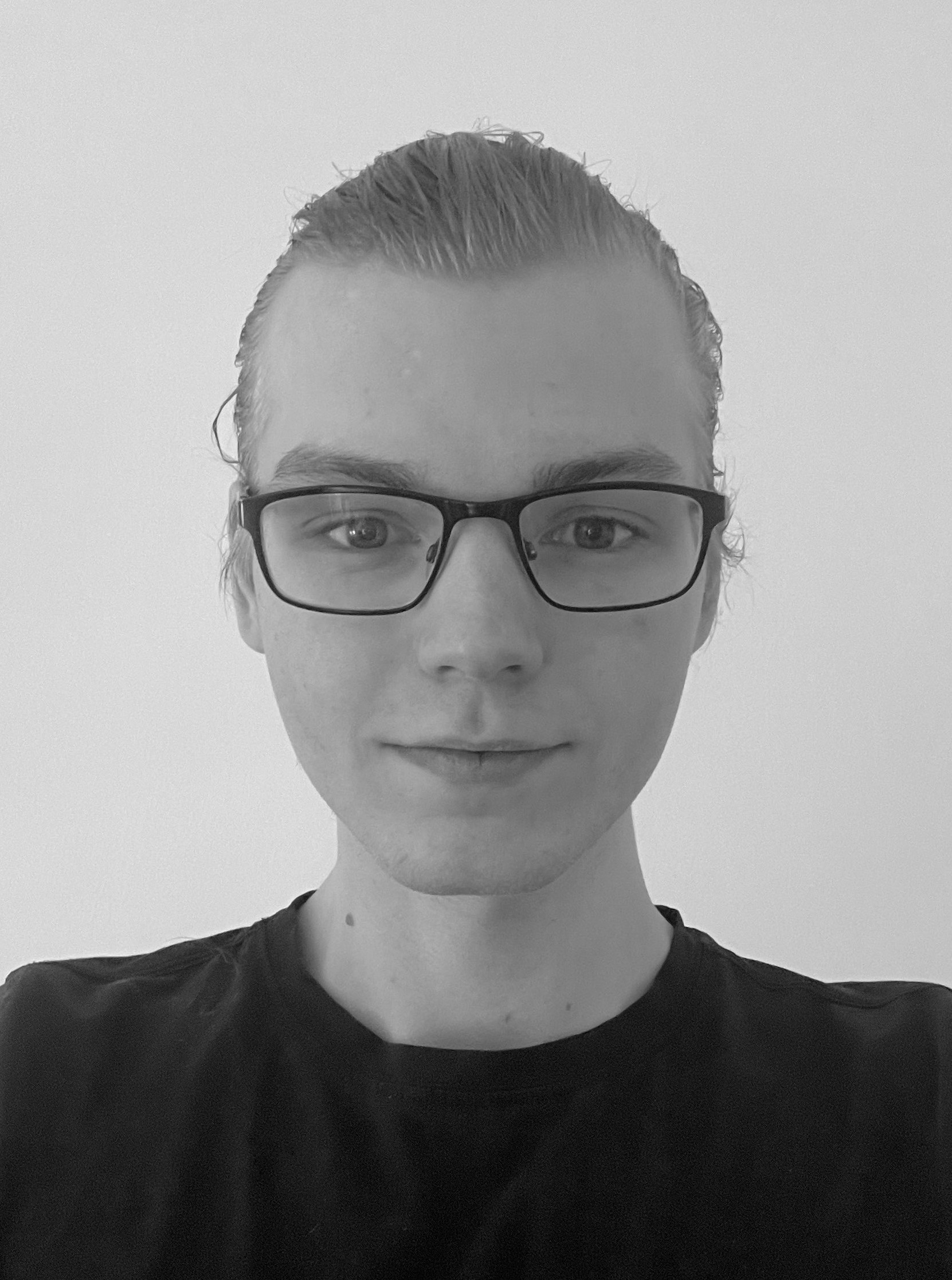}}]{Christopher Lindenberg} is currently pursuing his B.Sc. degree in Computer Science together with Steffen Seitz and Ronald Tetzlaff at the Technische Universität Dresden in Germany. He is recently researching neural network-based failure diagnosis and condition monitoring at the Chair of Fundamentals of Electrical Engineering. 
\end{IEEEbiography}

\begin{IEEEbiography}[{\includegraphics[width=1in,clip,keepaspectratio]{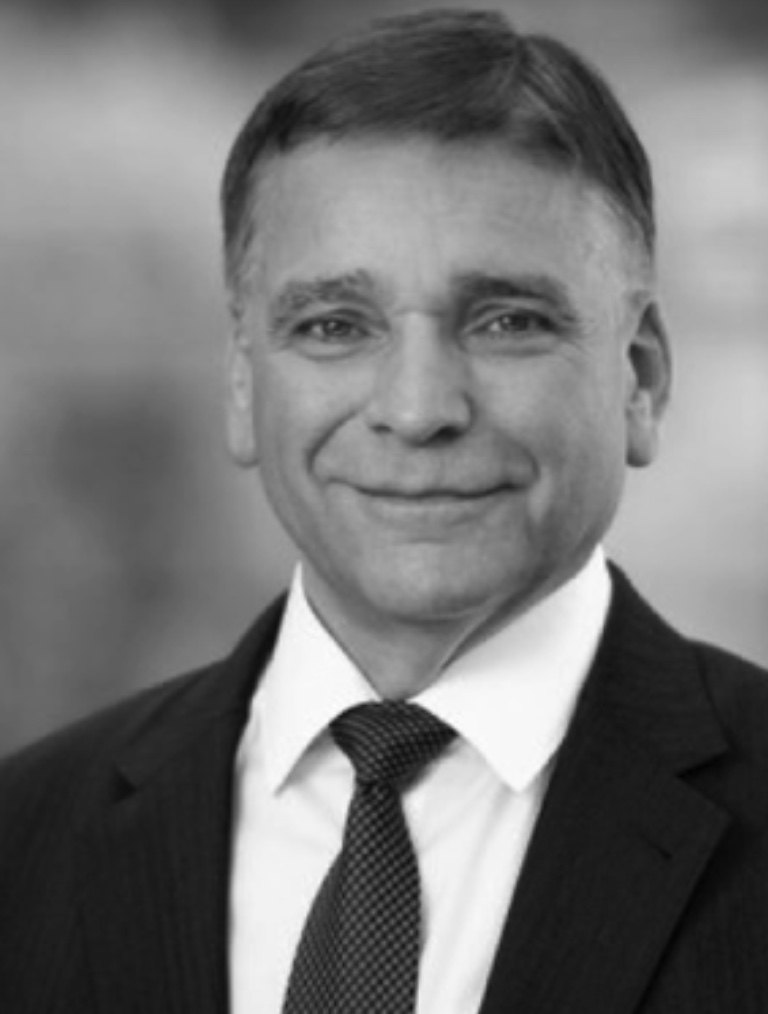}}]{Ronald Tetzlaff}
	(Senior Member, IEEE) is the Head of the Chair of Fundamentals of Electrical Engineering, Technische Universität Dresden, Germany. Since August 2020, he is also the Chief Officer of Technology Transfer and Internationalization at the Technische Universität Dresden. His scientific interests include problems in signal and system theory, medical signal processing, stochastic processes, system modeling, system identification, machine learning, mem-elements, memristive systems, Volterra systems, physical fluctuation phenomena, and Cellular Nonlinear Networks.  
\end{IEEEbiography}

\begin{IEEEbiography}[{\includegraphics[width=1in,clip,keepaspectratio]{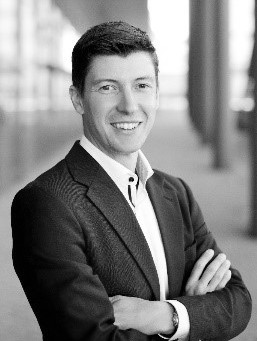}}]{Stephan Schlegel} received the doctoral and the habilitation degree in electrical engineering at the Technische Universität Dresden in 2011 and 2019. He led the research group “Electrical Contacts and Connection” between 2011 and 2020. Since 2020, he is provisional holder of the Chair of High Voltage and High Current Engineering at the TU Dresden. His research focus is in the fields of the fundamentals of the electrical-mechanical-thermal contact behavior as well as the aging mechanisms and long-term behavior of electrical contacts and connections.%\\\\\\\\\\\\\\\\\\\\\\\\
\end{IEEEbiography}

\end{document}